\documentclass{article}
\usepackage{cite}
\usepackage{amsmath,amssymb,amsfonts}

\usepackage{graphicx}
\usepackage{textcomp}
\usepackage{booktabs}       
\usepackage{multirow}       
\usepackage{graphicx}       
\usepackage{colortbl}       
\usepackage[table,xcdraw]{xcolor}  
\usepackage{amsmath}        
\usepackage{cite}           

\usepackage{longtable}      
\usepackage{adjustbox}      
\usepackage{algorithm}
\usepackage[noend]{algpseudocode}
\usepackage{cite}
\usepackage{arxiv}
\usepackage{amsmath,amssymb,amsfonts}
\usepackage[utf8]{inputenc} 
\usepackage[T1]{fontenc}    
\usepackage{hyperref}       
\usepackage{url}            
\usepackage{booktabs}       
\usepackage{amsfonts}       
\usepackage{nicefrac}       
\usepackage{microtype}      
\usepackage{lipsum}
\usepackage{graphicx}
\graphicspath{ {./images/} }
\usepackage{algorithm}

\title{DiffPose-Animal: A Language-Conditioned Diffusion Framework for Animal Pose Estimation}

\author{
  Tianyu Xiong\thanks{These authors contributed equally to this work.} \\
  Guanghua Cambridge International School\\
  Shanghai, China \\
  \texttt{17317233654@163.com} \\
  \And             
  Chang Gao\footnotemark[1] \\
  Shanghai World Foreign Language Academy, WFLA\\
  Shanghai, China \\
  \texttt{victorgao2025@126.com} \\
  \And
  Dayi Tan\thanks{Corresponding author: dytan@tongji.edu.cn} \\
  School of Automotive Studies\\
  Tongji University\\
  Shanghai, China \\
  \texttt{dytan@tongji.edu.cn} \\
  \And
  Wei Tian \\
  School of Automotive Studies\\
  Tongji University\\
  Shanghai, China \\
  \texttt{tian\_wei@tongji.edu.cn}
}

\begin{document}


\maketitle
\begin{abstract}
Animal pose estimation is a fundamental task in computer vision, with growing importance in ecological monitoring, behavioral analysis, and intelligent livestock management. Compared to human pose estimation, animal pose estimation is more challenging due to high interspecies morphological diversity, complex body structures, and limited annotated data. In this work, we introduce DiffPose-Animal, a novel diffusion-based framework for top-down animal pose estimation. Unlike traditional heatmap regression methods, DiffPose-Animal reformulates pose estimation as a denoising process under the generative framework of diffusion models. To enhance semantic guidance during keypoint generation, we leverage large language models (LLMs) to extract both global anatomical priors and local keypoint-wise semantics based on species-specific prompts. These textual priors are encoded and fused with image features via cross-attention modules to provide biologically meaningful constraints throughout the denoising process. Additionally, a diffusion-based keypoint decoder is designed to progressively refine pose predictions, improving robustness to occlusion and annotation sparsity. Extensive experiments on public animal pose datasets demonstrate the effectiveness and generalization capability of our method, especially under challenging scenarios with diverse species, cluttered backgrounds, and incomplete keypoints. Code will be available at https://github.com/cici203/Diffpose\_animal.

\end{abstract}


\section{Introduction}
\label{sec:introduction}

In the domains of ecological conservation and biological research, animal behavior plays a crucial role in environmental monitoring and ecosystem protection. Estimating animal poses allows for the analysis of natural behaviors, thereby enabling the evaluation of ecological health status \cite{li2023scarcenet}. Similarly, in intelligent surveillance systems, the detection of abnormal behaviors, such as recumbency or unconsciousness in target animals, contributes to animal safety monitoring and promotes the advancement of intelligent livestock management in agricultural settings \cite{wiltshire2023deepwild}. Furthermore, compared to humans, animals, particularly mammals, generally exhibit more complex three-dimensional structures and movement patterns. Since motion capture and modeling operations are fundamentally based on pose estimation, this technique holds substantial significance for applications including animal virtual reality and image reconstruction \cite{xu2023animal3d}.

In the field of computer vision, pose estimation is a specific task aimed at locating the joints or predefined keypoints of a target (e.g., the human body) in a given image or video, which can concisely and vividly reflect both the static and dynamic information of the target \cite{10.1145/3603618}, \cite{WANG2021103225}. It has been widely applied in tasks such as action recognition \cite{choutas2018potion}, object tracking \cite{andriluka2018posetrack}, and 3D reconstruction \cite{guler2019holopose}. Initially, human pose estimation dominated as the mainstream research focus. With the advancement of hardware computational resources, the release of large-scale datasets, and the rapid development of deep learning methods, research on pose estimation has experienced explosive growth and the variety of detected objects has become increasingly diverse. In particular, animals play a crucial role in ecosystems and their behavioral patterns are closely related to humans, making them a key focus in current pose estimation research \cite{lauer2022multi}, \cite{li2021synthetic}.

Although significant progress has been made in human pose estimation technology, animal pose estimation remains in its early developmental stages, with substantial challenges persisting in achieving accurate pose estimation. Several key factors contribute to these difficulties. First, human appearance exhibits relative consistency, characterized by smooth skin and clothing coverage, whereas animals demonstrate remarkable interspecies variation in surface characteristics. For instance, hippopotamuses possess smooth skin while tigers exhibit dense fur coverage. Second, human postures are relatively uniform, typically maintaining an upright standing position, whereas animals display complex and variable postures including crawling, lying, leaping, and other extreme body configurations. Additionally, animals frequently inhabit complex natural environments characterized by diverse backgrounds, frequent occlusions, truncation phenomena, and challenging lighting conditions. What is more, the extensive diversity among animal species also results in significant variance in data distribution across different species. These challenges, combined with the current limitations in scale and diversity of animal pose datasets, substantially increase the difficulty of achieving robust animal pose estimation in complex real-world scenarios.


In this work, we propose DiffPose-Animal, a novel diffusion-based framework for top-down animal pose estimation. Our method introduces denoising diffusion probabilistic modeling into the keypoint estimation pipeline, enhancing robustness to occlusions and morphological diversity often present in animal datasets. Moreover, to better guide the diffusion process, we incorporate semantic priors derived from large language models (LLMs), which provide both global species-level anatomical context and local keypoint-wise descriptions. These textual cues are fused with visual features through cross-attention mechanisms to improve structural plausibility and semantic alignment of predicted poses. The main contributions are summarized as follows:

\begin{itemize}
    \item We propose DiffPose-Animal, the first framework to introduce denoising diffusion probabilistic modeling into animal pose estimation, reformulating keypoint prediction as a generative denoising process to enhance robustness under occlusion, cluttered background, and limited annotations.
    
    \item We design a global-local semantic prompting strategy using large language models (LLMs), which extracts species-aware anatomical priors and keypoint-level descriptions. These textual cues are encoded and fused with image features through cross-attention for structurally coherent pose generation.
    
    \item We develop a multimodal fusion pipeline and a diffusion-driven pose decoder that iteratively refines the keypoint heatmaps, achieving strong generalization across species and superior performance on multiple public animal pose benchmarks.
\end{itemize}

\section{Related work}
\label{sec: Related work}

\subsection{Animal pose estimation}
Early research in animal pose estimation was primarily constrained by limited datasets, with most studies focusing on addressing data scarcity issues. Some approaches leveraged synthetic data through unsupervised or self-supervised learning paradigms. For instance, CC-SSL \cite{mu2020learning} employed CAD models to generate synthetic animal data with pseudo-labels, followed by self-supervised training based on these generated annotations. However, such synthetic data exhibited domain shifts in comparison to real-world data and contained inherent noise. To mitigate these limitations, UDA-Animal-Pose \cite{li2021synthetic} introduced a multi-scale domain adaptation module to reduce discrepancies between synthetic and real data, coupled with a coarse-to-fine pseudo-label refinement strategy. Further advancing this direction, ScarceNet \cite{li2023scarcenet} proposed a novel semi-supervised learning framework that enhanced pseudo-label reliability, effectively alleviating data scarcity challenges. Other works explored transfer learning and domain generalization strategies for animal pose estimation. For example, AnimalPose \cite{cao2019cross} developed a cross-domain adaptation method to transfer knowledge from labeled to unlabeled animal species while also leveraging human pose priors, demonstrating empirically validated effectiveness. DeepCut \cite{pishchulin2016deepcut}, building upon the open-source DeepLabCut \cite{mathis2018deeplabcut} framework, established a multi-task training architecture for simultaneous animal keypoint and limb prediction, further showing that ImageNet-pre-trained models could enhance both estimation accuracy and generalization capability. Nevertheless, these methods were typically limited to a narrow subset of species. Recent years have witnessed significant progress with the introduction of large-scale animal pose datasets such as AP-10K \cite{yu2021ap} and AnimalKingdom \cite{ng2022animal}. Capitalizing on these developments, AnimalRTPose \cite{WU2025107685} implements high-frequency feature extraction through its CSPNeXt \cite{chen2024cspnext} backbone network, integrates channel attention mechanisms to optimize high/low-frequency feature fusion, and employs spatial pyramid pooling for multi-scale contextual representation. This architecture enables robust feature encoding across varying spatial resolutions, substantially improving adaptability to morphologically diverse species and complex environments.

\subsection{Human pose estimation}
Human pose estimation is a fundamental task in computer vision, aiming to localize key anatomical landmarks such as the head, shoulders, elbows, and knees from visual input. Existing human keypoint estimation methods can be categorized into three paradigms: single-stage \cite{tan2024diffusionregpose}, bottom-up \cite{cheng2020higherhrnet}, and top-down \cite{xiao2018simple}. In the single-stage paradigm, the poses and bounding boxes of all targets in an image are predicted simultaneously. This paradigm presents significant challenges in prediction accuracy, often failing to meet the requirements of practical applications. Additionally, it typically requires complex models for feature extraction and keypoint decoding. Under the bottom-up method, the model predicts the keypoint locations of all target instances in the input image along with the association information between keypoints. Subsequently, post-processing methods such as grouping are employed to assign keypoints to their corresponding instances based on the association information. However, since the grouping process in this paradigm relies on heuristic methods involving numerous empirical techniques, its performance generally falls short of that of the top-down paradigm. With advancements in foundational research such as object detection \cite{chen2023diffusiondet}, the top-down approach has emerged as the dominant paradigm for pose estimation. In this paradigm, a detector first identifies the bounding boxes of target instances, which are then expanded by a certain ratio and cropped. Pose estimation is subsequently performed within these instance-specific bounding boxes. With respect to this, we primarily focus on the top-down paradigm in pose estimation. 

Typically, Stacked Hourglass Network~\cite{newell2016stacked} adopts a multi-stage symmetrical encoder-decoder architecture with skip connections to capture rich spatial and contextual information across multiple scales. By repeatedly downsampling and upsampling feature maps with skip connections, it enables iterative refinement of pose predictions and effectively models complex body structures. Subsequent methods such as the Cascaded Pyramid Network (CPN)~\cite{chen2018cascaded} enhance representation learning through multi-scale feature fusion and cascaded refinement, leading to more accurate joint localization. SimpleBaseline~\cite{xiao2018simple} streamlines the design by appending deconvolution layers to a ResNet~\cite{he2016deep} backbone, while HRNet~\cite{sun2019deep} maintains high-resolution representations throughout the network and achieves strong performance through continuous multi-scale feature fusion. To address the limitations of heatmap quantization, SimCC~\cite{li2022simcc} reformulates keypoint regression as two independent 1D classification tasks, enabling sub-pixel level localization accuracy. Transformer-based ViTPose~\cite{xu2022vitpose} further decouples feature encoding and keypoint prediction by employing a pure Vision Transformer backbone, and supports both heatmap and coordinate classification heads for greater modeling flexibility. In addition to architectural innovations, recent work has focused on improving the quality of supervision. RLE~\cite{li2021human} loss models more realistic keypoint distributions as soft targets, mitigating the bias introduced by idealized Gaussian heatmaps. Despite impressive accuracy, top-down methods often suffer from high computational cost due to redundant bounding boxes and large pose variability. To address these issues, DynPose~\cite{xu2025dynpose} introduces a dynamic inference framework that employs a lightweight router to dispatch pose instances to either a small or large pose estimator based on complexity, achieving a more favorable balance between accuracy and efficiency.


\subsection{Diffusion model}

Diffusion models have become a dominant paradigm in generative modeling due to their ability to synthesize high-fidelity data. Denoising Diffusion Probabilistic Models (DDPMs)~\cite{ho2020denoising} introduce a Markovian forward process that gradually adds noise to data and a reverse process learned to recover it, yielding impressive results across image and audio domains. However, their slow sampling speed has prompted various improvements. DDIM~\cite{song2020denoising} proposes a non-Markovian deterministic formulation that accelerates inference while preserving generation quality. To improve scalability, Latent Diffusion Models (LDMs)~\cite{rombach2022high} perform diffusion in a compressed latent space, significantly reducing computation and enabling high-resolution image generation. Meanwhile, Diffusion Transformers (DiTs)~\cite{peebles2023scalable} replace U-Nets~\cite{ronneberger2015u} with transformer-based backbones, offering superior modeling of long-range dependencies and strong performance on large-scale benchmarks. Beyond architectural innovations, recent developments such as Score-based Generative Models (SGMs)~\cite{song2020score} and Consistency Models~\cite{song2023consistency} further enhance efficiency and theoretical grounding. These advances collectively establish diffusion models as a flexible and scalable foundation for diverse generative tasks.

\section{Method}
\label{sec:Method}


\subsection{Preliminary}
\label{sec:preliminary}

Denoising diffusion probabilistic models (DDPMs)~\cite{ho2020denoising} are a class of generative models that learn to synthesize data through a gradual denoising process. They consist of two main phases: a forward diffusion process and a reverse denoising process. The framework has demonstrated strong performance in modeling complex data distributions and is increasingly applied to structured prediction tasks, such as image/video synthesis \cite{ho2022video}, audio processing \cite{choi2023diffv2s}, view synthesis \cite{chen2023single} and human pose estimation~\cite{tan2024diffusionregpose}.

\subsubsection*{Forward Diffusion Process}

Given a data sample $\mathbf{x}_0 \sim q(\mathbf{x})$, the forward process gradually corrupts $\mathbf{x}_0$ into a sequence of latent variables $\{\mathbf{x}_t\}_{t=1}^T$ over $T$ steps by adding Gaussian noise:

\begin{equation}
q(\mathbf{x}_t | \mathbf{x}_{t-1}) = \mathcal{N}(\mathbf{x}_t; \sqrt{1 - \beta_t} \, \mathbf{x}_{t-1}, \beta_t \mathbf{I}),
\end{equation}

where $\beta_t \in (0, 1)$ is a variance schedule.To simplify the sampling process, one can directly compute $\mathbf{x}_t$ from $\mathbf{x}_0$ through a deterministic formula:

\begin{equation}
q(\mathbf{x}_t | \mathbf{x}_0) = \mathcal{N}(\mathbf{x}_t; \sqrt{\bar{\alpha}_t} \, \mathbf{x}_0, (1 - \bar{\alpha}_t) \mathbf{I}),
\end{equation}

where the following notations are defined:

\begin{equation}
\alpha_t = 1 - \beta_t, \quad \bar{\alpha}_t = \prod_{s=1}^t \alpha_s.
\end{equation}

Here, $\bar{\alpha}_t$ represents the cumulative product of retained information up to step $t$, and $(1 - \bar{\alpha}_t)$ corresponds to the aggregated noise variance.

\subsubsection*{Reverse Denoising Process}

The goal of the reverse process is to learn a parameterized model $p_\theta(\mathbf{x}_{t-1}|\mathbf{x}_t)$ that gradually removes noise from $\mathbf{x}_T$ to recover a clean data sample. In practice, this is typically implemented by training a neural network $\epsilon_\theta$ to predict the noise $\boldsymbol{\epsilon}$ added in the forward process:

\begin{equation}
\mathcal{L}_{\text{simple}} = \mathbb{E}_{\mathbf{x}_0, \boldsymbol{\epsilon}, t} \left[ \left\| \boldsymbol{\epsilon} - \epsilon_\theta(\mathbf{x}_t, t) \right\|_2^2 \right],
\end{equation}

where $\mathbf{x}_t = \sqrt{\bar{\alpha}_t} \, \mathbf{x}_0 + \sqrt{1 - \bar{\alpha}_t} \, \boldsymbol{\epsilon}$ and $\boldsymbol{\epsilon} \sim \mathcal{N}(0, \mathbf{I})$. This training objective enables the model to learn a denoising function that progressively estimates the clean signal from noisy observations.

\subsection{DiffPose-Animal}

\begin{figure*}[tbp]\centering
 \includegraphics[width=\linewidth]{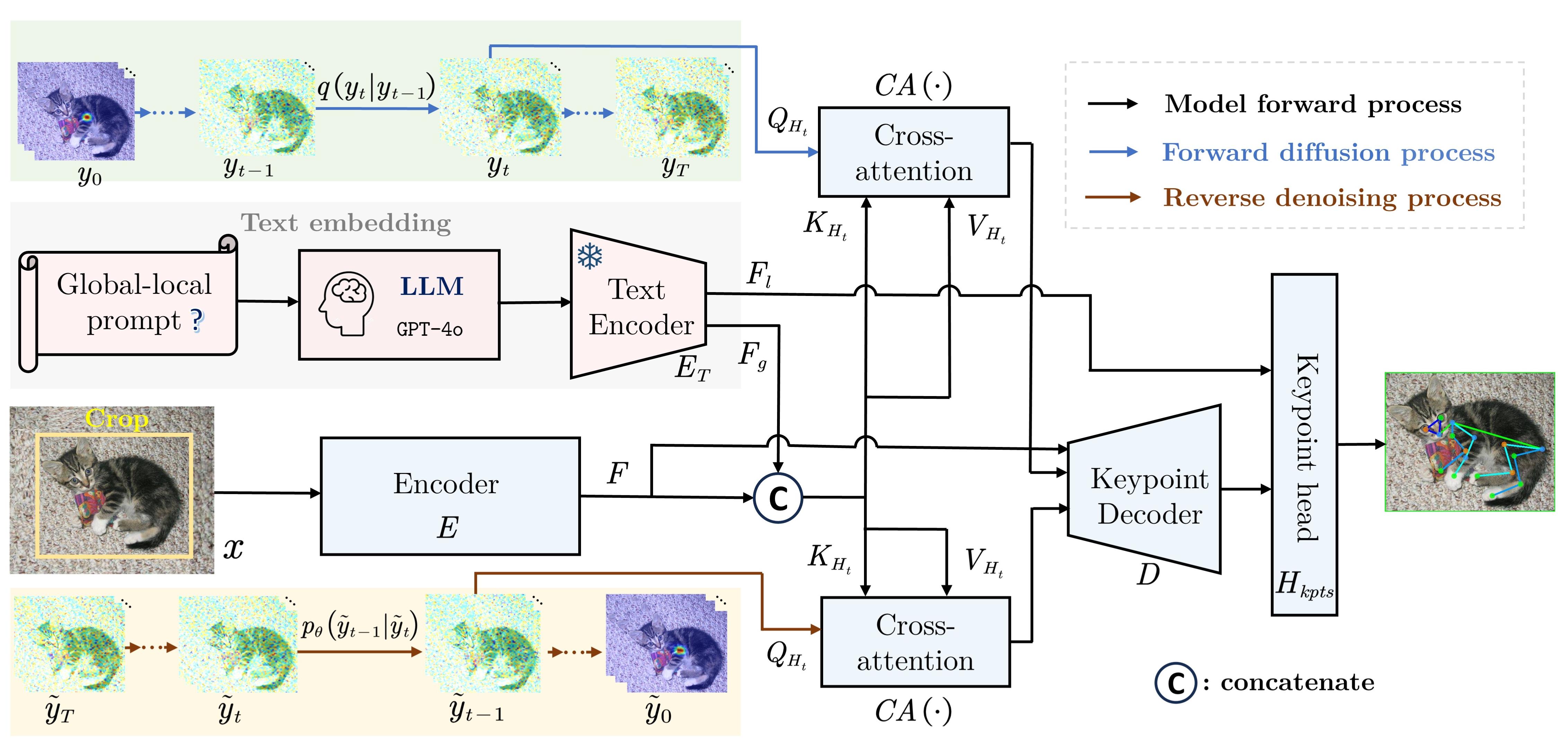}
\caption{
    The proposed DiffPose-Animal framework. Animal pose estimation is modeled as a conditional denoising process, guided by semantic priors from LLMs and visual features through cross-attention-based multimodal fusion. Given a cropped animal image, global and keypoint-specific anatomical prompts are fed into an LLM to generate textual descriptions, which are encoded into global ($F_g$) and local ($F_l$) semantic priors. These are fused with visual features $F$ from a vision encoder. A diffusion-based decoder then iteratively denoises an initial noisy pose $\tilde{y}_T$ to predict the final keypoints $\tilde{y}_0$, guided by cross-attention with the multimodal context at each timestep.
}
\label{fig:framework}
\end{figure*}

\subsubsection{Architecture}

We propose DiffPose-Animal, a diffusion-based framework for animal pose estimation that integrates semantic anatomical priors from large language models (LLMs). By leveraging global-local prompts, the model encodes species- and keypoint-specific descriptions via a frozen text encoder and fuses them with visual features through a cross-attention-guided denoising process. The multimodal representation guides the reverse diffusion to iteratively refine keypoint predictions, enabling robust and anatomically informed pose estimation.

\textbf{Image encoder.} To extract multi-scale visual features from raw animal images, we employ a convolutional backbone as the image encoder. Specifically, we utilize HRNet \cite{sun2019deep} as the backbone due to its effectiveness in preserving spatial resolution and capturing rich semantic information. The extracted hierarchical feature maps serve as the visual input to the multimodal fusion module, where they are integrated with the global textual embeddings generated by the LLM.

\textbf{Text embedding module.} 
To integrate anatomical priors into the pose estimation process, we design a text embedding module that extracts both global and local semantic features from LLMs. For each animal species, we craft a structured prompt that includes the list of keypoints and requests the LLM (e.g., GPT-4o \cite{hurst2024gpt}) to perform two tasks: (1) generate a global description covering the animal’s body structure, locomotion style, and biomechanical characteristics relevant to pose, and (2) provide short semantic descriptions for each keypoint, highlighting their anatomical roles or functions. 

The combined textual output is then encoded using a frozen text encoder (e.g., CLIP~\cite{radford2021learning}, BERT~\cite{reimers2019sentence}, or all-MiniLM-L6-v2 \cite{yin2024study} from SentenceTransformers), producing two types of embeddings: a global feature vector $F_g$ representing species-level anatomical context, and a set of local keypoint embeddings $F_l \in \mathbb{R}^{N \times d}$ where $N$ is the number of keypoints and $d$ is the embedding dimension. These features are later fused with the visual backbone outputs and heatmap representations to guide the pose generation process with semantically meaningful priors.

\textbf{Cross attention.} Cross-attention is employed to enhance semantic alignment between visual features and language-derived anatomical priors. Specifically, at each diffusion timestep $t$, the intermediate heatmap feature $Q_{H_t}$ serves as the query. The keys and values are constructed by concatenating the visual feature $F$ with the global textual embedding $F_g$, forming a multimodal representation that encodes both image appearance and anatomical context. This cross-attention mechanism enables the decoder to selectively attend to relevant semantic cues during the denoising process, facilitating more accurate keypoint localization.

\textbf{Keypoint decoder and keypoint head.} The keypoint decoder takes the output of the cross-attention module as its primary input and performs an element-wise addition with the visual feature map $F$ from the image encoder, forming a residual connection that preserves low-level visual cues. This residual-enhanced representation is subsequently fused with the local textual embeddings $F_l$, which encode fine-grained anatomical meanings of individual keypoints. The resulting multimodal feature is used to generate a set of heatmaps corresponding to each anatomical keypoint, which are then decoded to produce the final animal pose estimation.

\subsubsection{Training}
During training (Blue flow + Black flow in Fig. \ref{fig:framework}), we first construct a forward diffusion process by progressively adding Gaussian noise to the ground-truth heatmaps, resulting in a series of noisy heatmap representations. The model is then trained to reverse this diffusion trajectory and recover the original heatmap through denoising. Algorithm~\ref{alg:diffpose_training} presents the pseudo-code for the training procedure of our proposed DiffPose-Animal framework.

Specifically, at each training iteration, a timestep $t$ is randomly sampled, and the corresponding noisy heatmap $\mathbf{H}_t$ is used as the query $Q_{H_t}$. Simultaneously, the input image is passed through the image encoder to extract visual features $\mathbf{F}$, which are concatenated with the textual embeddings from the Text Embedding Module to form the key $K_{H_t}$ and value $V_{H_t}$ inputs for the cross-attention module. The output of the cross-attention is fused with $\mathbf{F}$ via a residual connection and fed into the keypoint decoder.

The decoder predicts heatmaps for all keypoints, which are supervised using a Mean Squared Error (MSE) loss against the ground-truth heatmaps. This diffusion-based training strategy allows the model to learn robust keypoint representations under varying levels of noise.

\begin{center}
\begin{minipage}{0.8\linewidth}  
\begin{algorithm}[H]
\caption{Training DiffPose-Animal}
\label{alg:diffpose_training}
\begin{algorithmic}[1]
\Require Image $x$, GT heatmap $y_0$, animal category $c$, keypoint list $K$, total diffusion steps $T$, pretrained LLM, model $f_\theta$ with encoder $E$, text encoder $E_T$, keypoint head $H_{kpts}$
\While{not converged}
    \State Sample timestep $t \sim \mathcal{U}(\{1, \ldots, T\})$
    \State Extract visual feature: $F \gets E(x)$
    \State Generate global-local prompt $\rightarrow$ text description
    \State Encode text: $(F_g, F_l) \gets E_T(\text{text})$
    \State Add noise to GT heatmap: $y_t \gets q(y_0, t)$
    \State Concatenate features: $F_{\text{fuse}} \gets [F; F_g]$
    \State Set query $Q_{H_t} \gets y_t$, keys/values $K_{H_t}, V_{H_t} \gets F_{\text{fuse}}$
    \State Cross-attention: $F_{CA} \gets \text{CA}(Q_{H_t}, K_{H_t}, V_{H_t})$
    \State Residual fusion: $F_D \gets F_{CA} + F$
    \State Keypoint decoding: $\hat{y}_0 \gets H_{kpts}(F_D, F_l)$
    \State Compute loss: $L \gets \| \hat{y}_0 - y_0 \|^2$
    \State Update model $f_\theta$ via gradient step $\nabla_\theta L$
\EndWhile
\end{algorithmic}
\end{algorithm}
\end{minipage}
\end{center}

\begin{center}
\begin{minipage}{0.8\linewidth}  
\begin{algorithm}[H]
\caption{Inference of DiffPose-Animal in $T$ iterative steps}
\label{alg:diffpose_inference}
\begin{algorithmic}[1]
\Require Image $x$, animal category $c$, keypoint list $K$, diffusion steps $T$, trained model $f_\theta$ with encoder $E$, text encoder $E_T$, keypoint head $H_{kpts}$
\State Extract visual feature: $F \gets E(x)$
\State Generate global-local prompt $\rightarrow$ text description
\State Encode text: $(F_g, F_l) \gets E_T(\text{text})$
\State Concatenate features: $F_{\text{fuse}} \gets [F; F_g]$
\State Initialize noisy heatmap: $\hat{y}_T \sim \mathcal{N}(0, \mathbf{I})$
\For{$t = T, T-1, \ldots, 1$}
    \State Set query $Q_{H_t} \gets \hat{y}_t$, keys/values $K_{H_t}, V_{H_t} \gets F_{\text{fuse}}$
    \State Cross-attention: $F_{CA} \gets \text{CA}(Q_{H_t}, K_{H_t}, V_{H_t})$
    \State Residual fusion: $F_D \gets F_{CA} + F$
    \State Keypoint decoding: $\hat{y}_{t-1} \gets H_{kpts}(F_D, F_l)$
\EndFor
\State \Return Final keypoint prediction: $\hat{y}_0$
\end{algorithmic}
\end{algorithm}
\end{minipage}
\end{center}

\subsubsection{Inference}

During inference (Brown flow + Black flow in Fig. \ref{fig:framework}), DiffPose-Animal first extracts visual features \(F\) from the input animal image \(x\) using an image encoder \(E\). In parallel, anatomical knowledge is generated via a pretrained large language model (LLM) based on a global-local prompt specific to the animal species and its keypoints. The resulting semantic descriptions are processed by a frozen text encoder \(E_T\), yielding global textual features \(F_g\) and local keypoint-wise features \(F_l\).

The global text embedding \(F_g\) is concatenated with the image feature \(F\) to form a fused multimodal representation \(F_{\text{fuse}} = [F; F_g]\). Simultaneously, the model samples an initial heatmap \(y_T\) from a Gaussian distribution to simulate noisy keypoint predictions. It then performs an iterative reverse diffusion process, denoising the heatmap from \(y_T\) to the final prediction \(\hat{y}_0\).

At each denoising step \(t\), the noisy heatmap \(y_t\) serves as the query \(Q_{H_t}\), while the multimodal features \(F_{\text{fuse}}\) are used to compute keys and values \((K_{H_t}, V_{H_t})\) for the cross-attention module \(CA(\cdot)\). The output \(F_{CA} = CA(Q_{H_t}, K_{H_t}, V_{H_t})\) is then fused with the visual features \(F\) via a residual connection to obtain \(F_D = F_{CA} + F\).

Finally, the refined feature \(F_D\) is fused with the local textual features \(F_l\), and passed through the keypoint decoder \(H_{\text{kpts}}\) to generate the predicted keypoint heatmaps \(\hat{y}_0\). A full description of the inference process is provided in Algorithm~\ref{alg:diffpose_inference}.

\section{Experiment}
\subsection{Datasets}

The \textbf{AP-10K \cite{yu2021ap}} dataset is one of the most diverse large-scale benchmarks for animal pose estimation. It contains 10,015 images across 23 animal species, with each instance annotated with 17 keypoints following a unified anatomical scheme. The dataset provides three different train-validation-test splits. AP-10K covers both domestic and wild animals in varied real-world scenarios, posing challenges such as scale variation, background clutter, and occlusion. It is particularly well-suited for evaluating cross-species generalization performance.

\textbf{AnimalPose \cite{cao2019cross}} is a benchmark dataset introduced for cross-species animal pose estimation. It contains approximately 5,517 images from five quadruped animal categories: dog, cat, horse, cow, and sheep. Each instance is annotated with 18 keypoints, corresponding to semantically aligned body joints across species. The dataset is relatively small in scale but provides high-quality keypoint annotations, making it suitable for evaluating models in low-data regimes and studying generalization across similar quadrupeds.

\textbf{Animal Kingdom \cite{ng2022animal}} is a recently introduced, large-scale and fine-grained animal pose dataset comprising over 24,000 images from an extensive taxonomy of 50+ species, spanning mammals, birds, and reptiles. Keypoint annotations are species-specific and vary in number, and the dataset includes both common and rare animals under natural environments. Due to its size, species diversity, and scene complexity, AnimalKingdom serves as a rigorous benchmark for evaluating the scalability and robustness of animal pose estimation methods under long-tailed and open-world settings.

\subsection{Evaluation Metrics}
\label{sec:metrics}

To comprehensively evaluate the performance of animal pose estimation models, we adopt a combination of standard keypoint evaluation protocols commonly used in both human and animal pose estimation literature, including the COCO evaluation protocol~\cite{lin2014microsoft}, Object Keypoint Similarity (OKS), Percentage of Correct Keypoints (PCK), and Area Under Curve (AUC).

\subsubsection{COCO Evaluation Protocol.}
Following the COCO benchmark, we use Average Precision (AP) and Average Recall (AR) as the primary metrics. The COCO-style AP is calculated by computing the mean of keypoint OKS over multiple thresholds ranging from 0.50 to 0.95 with a step size of 0.05:
\begin{itemize}
    \item \textbf{AP@0.50:} AP at OKS threshold = 0.50 (loosest match).
    \item \textbf{AP@0.75:} AP at OKS threshold = 0.75 (stricter match).
    \item \textbf{AP (mean):} Averaged over OKS thresholds from 0.50 to 0.95.
    \item \textbf{AR:} Average Recall over the same OKS thresholds.
\end{itemize}

OKS is used to measure the similarity between predicted and ground-truth keypoints, analogous to IoU in object detection. It accounts for object scale and keypoint annotation uncertainty. The OKS is defined as:

\begin{equation}
\text{OKS} = \frac{\sum_i \exp \left( - \frac{d_i^2}{2s^2k_i^2} \right) \cdot \delta(v_i > 0)}{\sum_i \delta(v_i > 0)},
\end{equation}

where:
\begin{itemize}
    \item $d_i$ is the Euclidean distance between the $i$-th predicted and ground-truth keypoints.
    \item $s$ is the object scale (typically the square root of the object’s bounding box area).
    \item $k_i$ is a per-keypoint constant controlling the falloff, reflecting the annotation uncertainty.
    \item $v_i$ is the visibility flag ($0$: not labeled, $1$: labeled but not visible, $2$: visible).
    \item $\delta(v_i > 0)$ is an indicator function to consider only labeled keypoints.
\end{itemize}

 \subsubsection{Percentage of Correct Keypoints (PCK).}
PCK measures the percentage of correctly predicted keypoints whose distance to the ground truth is within a given normalized threshold. A prediction is considered correct if:

\begin{equation}
\frac{\| \hat{\boldsymbol{x}}_i - \boldsymbol{x}_i \|_2}{\text{norm}} < \alpha,
\end{equation}

where $\hat{\boldsymbol{x}}_i$ and $\boldsymbol{x}_i$ denote the predicted and ground-truth coordinates respectively, and $\text{norm}$ is typically a function of object scale (e.g., torso size or bounding box size). We report PCK@0.05, which uses a threshold $\alpha = 0.05$.

\subsubsection{Area Under the Curve (AUC).}
AUC evaluates pose estimation performance over a continuous range of thresholds by plotting the PCK curve from $\alpha = 0$ to $\alpha = 0.5$, then computing the area under this curve. It provides a more robust measure of keypoint localization accuracy across multiple tolerance levels.

\subsection{Implementation details}
During the training phase, we employ the AdamW optimizer with a weight decay of $1 \times 10^{-4}$ and train the model for 210 epochs. The training is conducted on three NVIDIA RTX 3090 GPUs with a batch size of 48. The initial learning rate is set to $5 \times 10^{-4}$ and is decayed by a factor of 0.1 at the 170\textsuperscript{th} and 200\textsuperscript{th} epochs.

\subsection{Performance Comparison Across Multiple Datasets}

\begin{figure*}[ht]\centering
 \includegraphics[width=0.9\linewidth]{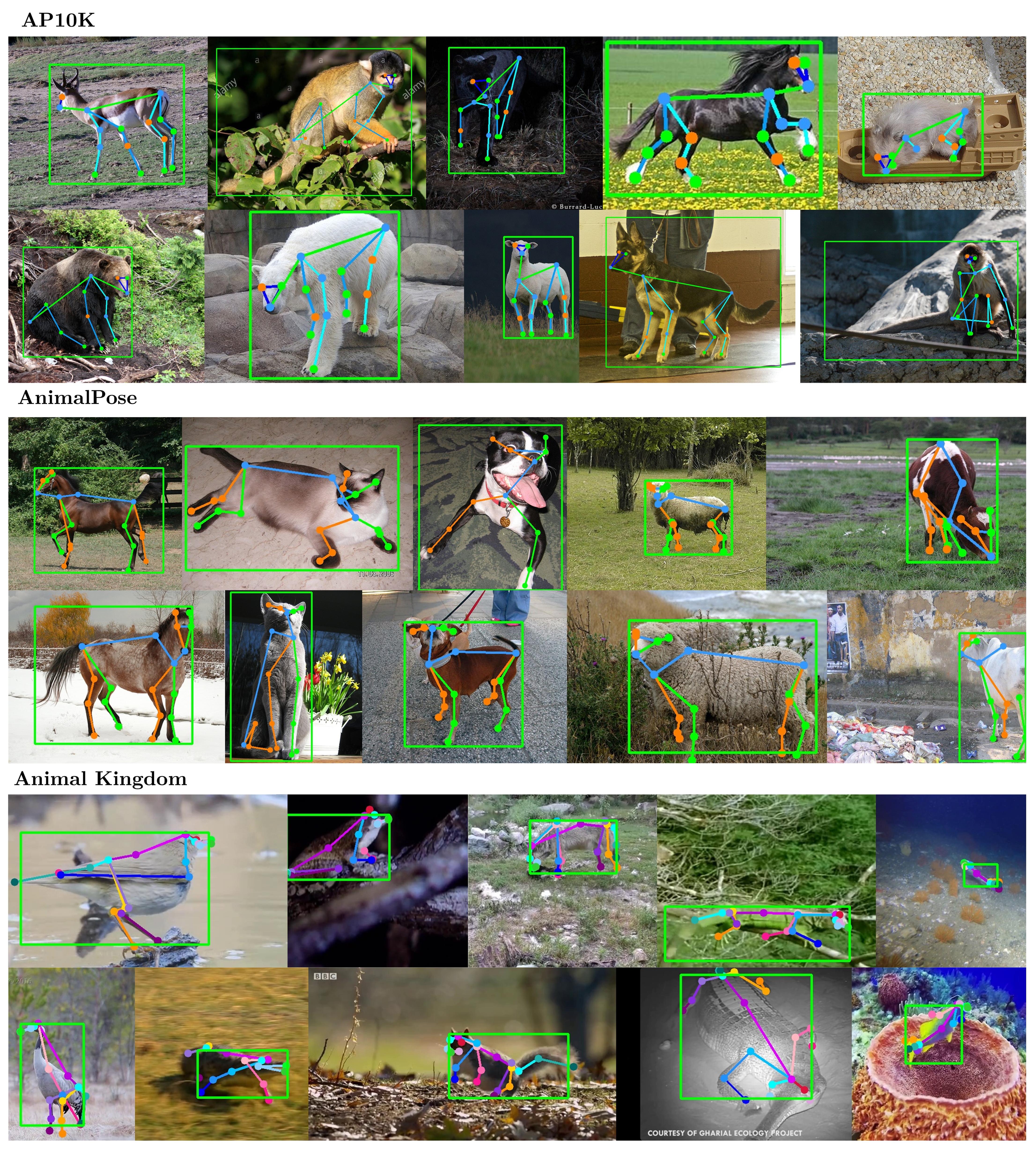}
\caption{
    Visualization results of DiffPose-Animal. From top to bottom: AP10K \cite{yu2021ap}, AnimalPose \cite{cao2019cross}, and AnimalKingdom \cite{ng2022animal}. These examples showcase the diversity of animal categories, poses, environments, and keypoint configurations across datasets. AP-10K includes various mammals under natural conditions; AnimalPose mainly contains domestic animals such as dogs and horses in relatively constrained scenes; Animal Kingdom spans broader species including birds, amphibians,  mammal, reptiles and marine animals, highlighting the challenge of cross-species generalization. 
}
\label{fig:vis_ap10k}
\end{figure*}

\begin{table*}
    \centering
    \tiny
    
    \caption{Comparisons with state-of-the-art methods on AP-10K.  Best values are in bold. ``-'' denotes unavailable values.}
    \label{Tab_hpe}
    \resizebox{0.95\textwidth}{!}{
    \begin{tabular}{cccccccc}
        \hline
        Method & Backbone &  AP & AP$_{50}$ & AP$_{75}$ & AP$_M$ & AP$_L$ & AR \\
        \hline

         \multirow{5}*{YOLOv8-Pose \cite{jocher2023ultralytics}} & CSPDarknet-N      & 45.90 &- & - & - & -&57.50  \\

             & CSPDarknet-S      & 51.00 &- & - & - & -&62.10  \\

             & CSPDarknet-M      & 56.40 &- & - & - & -&66.90  \\

             & CSPDarknet-L      & 59.70 &- & - & - & -&69.40  \\

             & CSPDarknet-X      & 61.00 &- & - & - & -&70.20  \\
            \hline

            AnimalRTPose-N \cite{WU2025107685} & CSPNeXt$^{\dagger}$-N
      & 45.30 &- & - & - & -&58.50  \\
            
            AnimalRTPose-S \cite{WU2025107685} & CSPNeXt$^{\dagger}$-S      & 50.80 &- & - & - & -&63.40  \\
            
            AnimalRTPose-M \cite{WU2025107685} & CSPNeXt$^{\dagger}$-M     & 56.60 &- & - & - & -&69.00  \\

            AnimalRTPose-L \cite{WU2025107685} & CSPNeXt$^{\dagger}$-L      & 62.00 &- & - & - & -&71.90  \\

            AnimalRTPose-X \cite{WU2025107685} & CSPNeXt$^{\dagger}$-X     & 65.10 &- & - & - & -&74.40 \\
            \hline

            Rtmpose \cite{lu2024rtmo} & Rtmpose-m      & 72.20 &93.90 & 78.80 & 56.90 & 72.80&- \\
            \hline

             DiffPose-Animal-split1& \multirow{4}*{HRNet-W32}  & 73.91 & \textbf{95.82} & \textbf{81.40} & 52.91 & 74.24 & \textbf{77.13} \\

             DiffPose-Animal-split2&   & 73.91 & 95.74 & 80.83 & \textbf{66.37}& \textbf{74.35} & \textbf{77.13} \\

             DiffPose-Animal-split3&  & \textbf{74.08} & 95.10 & 80.68 & 61.11& 74.30 & 77.04 \\

             DiffPose-Animal-avg&   & 73.97 & 95.55 & 81.03 & 60.13& 74.30 & 77.10 \\

        \hline
    \end{tabular}
    }
\end{table*}

\noindent
\textbf{Experimental Results on AP-10K.}
Table~\ref{Tab_hpe} provides a systematic comparison of state-of-the-art methods on the AP-10K dataset, illustrating performance scaling with backbone capacity, as supported by that both YOLOv8-Pose and AnimalRTPose exhibit monotonic performance gains when transitioning from lightweight to large backbones. In details, YOLOv8-Pose improves from 45.90 AP (CSPDarknet-N) to 61.00 AP (CSPDarknet-X), with a corresponding AR increase from 57.50 to 70.20. Moreover, AnimalRTPose, built on the more advanced CSPNeXt$^\dagger$ backbone, exhibits superior results in the same scale comparison, with AP ranging from 45.30 to 65.10 as CSPNeXt$^\dagger$-N changed to CSPNeXt$^\dagger$-X, highlighting the benefits of enhanced feature representation and stronger inductive biases. In contrast to these lightweight real-time architectures, RTMPose-m establishes a robust top-down benchmark with significantly higher metrics (72.20 AP and 93.90 AP$_{50}$), attributable to its sophisticated pose-specific head architecture. However, this approach remains limited by the absence of explicit semantic guidance and global anatomical reasoning mechanisms. Moreover, our proposed DiffPose-Animal achieves significant performance improvements through two critical innovations: diffusion-based iterative keypoint refinement and multimodal anatomical priors derived from LLMs. Equipped with only a moderate-capacity HRNet-W32 backbone, the method establishes new state-of-the-art results across all evaluation metrics, achieving 73.97 AP on average across three evaluation splits, along with 95.55 AP$_{50}$, 81.03 AP$_{75}$, and 77.10 AR. Moreover, it maintains robust performance across medium and large animal scales (AP$_M$ = 60.13, AP$_L$ = 74.30), outperforming all baseline methods by a notable margin. These results validate the effectiveness of integrating global-local language priors and iterative denoising for precise and semantically aligned animal pose estimation. 

\begin{table*}[ht]
    \centering
    \tiny
    \caption{Comparisons with state-of-the-art methods on Animalpose. Best values are in bold, and ``-'' denotes unavailable values.}
    \label{Tab_hpe2}
    \resizebox{0.95\textwidth}{!}{
    \begin{tabular}{cccccccc}
        \hline
        Method & Backbone &  AP & $\mathrm{AP}_{50}$ & $\mathrm{AP}_{75}$ & $\mathrm{AP}_{M}$ & $\mathrm{AP}_{L}$ & AR \\
        \hline

            SimpleBaseline \cite{xiao2018simple} & ShuffleNetV1      & 67.20 &- & - & - & -&74.80  \\
            \hline

            YOLOv8-Pose \cite{jocher2023ultralytics} & CSPDarknet-N      & 69.50 &- & - & - & -&77.40  \\
            \hline

            AnimalRTPose \cite{WU2025107685} & CSPNeXt-N      & 72.00 &- & - & - & -&\textbf{80.00}  \\
            \hline

             HRNet \cite{sun2019deep} & HRNet-W32  & 73.20 & 94.78 & 81.08 & 70.75& 74.65 & 76.77 \\
             \hline

             DiffPose-Animal& HRNet-W32  & \textbf{73.77} & \textbf{95.90} & \textbf{81.10} & \textbf{71.77}& \textbf{74.96} & 77.28 \\

        \hline
    \end{tabular}
    }
\end{table*}

\textbf{Experimental Results on AnimalPose.} Table~\ref{Tab_hpe2} provides a systematic comparison of pose estimation methods on the Animalpose dataset, demonstrating clear evolutionary progress from early lightweight architectures to contemporary sophisticated designs. The performance trajectory begins with SimpleBaseline (ShuffleNetV1 backbone) establishing a lightweight benchmark at 67.20 AP and 74.80 AR. Subsequent improvements emerge with YOLOv8-Pose (CSPDarknet-N), which achieves 69.50 AP and 77.40 AR through its detection-optimized feature extraction. Subsequently, a more significant advancement comes with AnimalRTPose (CSPNeXt-N backbone), attaining 72.00 AP and 80.00 AR, underscoring the advantages of modern backbone architectures. Moving to top-down approaches, HRNet-W32 achieves 73.20 AP, 94.78 AP$_{50}$, and 81.08 AP$_{75}$, with strong performance on both medium and large-scale animals. As compared, our proposed DiffPose-Animal, while employing the same HRNet-W32 backbone, surpasses all existing methods by significant margins, achieving state-of-the-art results: 73.77 AP (+0.57 over HRNet), 95.90 AP${50}$, and 81.10 AP${75}$. Notably, it demonstrates particularly strong performance on medium (71.77 AP$_M$) and large (74.96 AP$_L$) animals, with an overall AR of 77.28. These consistent improvements validate the efficacy of our integrated approach combining anatomical knowledge priors with diffusion-based refinement, which enables more precise and robust pose estimation without requiring backbone architecture modifications.

\begin{table}
    \centering
    \tiny
    \caption{Comparisons with state-of-the-art methods on Animal Kingdom.  Best values are in bold. ``-'' denotes unavailable values.}
    \label{Tab_hpe3}
    \resizebox{0.7\textwidth}{!}{
    \begin{tabular}{ccccc}
        \hline
        Config & Method & Backbone &  PCK(0.05) & AUC  \\
        \hline

        \multirow{2}*{P1} & hrnet-w32  \cite{sun2019deep}& \multirow{2}*{HRNet-W32}  & 63.23 & - \\

         & DiffPose-Animal&   & \textbf{63.25} & 61.37 \\
        \hline

        \multirow{2}*{P2} & hrnet-w32  \cite{sun2019deep}& \multirow{2}*{HRNet-W32}  & 37.41 & - \\

         & DiffPose-Animal&   & \textbf{38.24} & 39.82 \\
        \hline

        \multirow{2}*{P3-mammals} & hrnet-w32  \cite{sun2019deep}& \multirow{2}*{HRNet-W32}  & 57.10 & - \\

        & DiffPose-Animal&   & \textbf{58.36} & 60.11 \\
        \hline

        \multirow{2}*{P3-amphibians}  & hrnet-w32  \cite{sun2019deep}& \multirow{2}*{HRNet-W32}  & 53.58 & - \\

         & DiffPose-Animal&   & \textbf{54.08} & 52.17 \\
        \hline

        \multirow{2}*{P3-reptiles} & hrnet-w32  \cite{sun2019deep}& \multirow{2}*{HRNet-W32}  & 51.00 & - \\

        & DiffPose-Animal&   & \textbf{51.86} & 45.57\\
        \hline

         \multirow{2}*{P3-birds} & hrnet-w32  \cite{sun2019deep}& \multirow{2}*{HRNet-W32}  & 76.71 & - \\

         & DiffPose-Animal&   & \textbf{76.91} & 73.66\\
        \hline

         \multirow{2}*{P3-fishes} & hrnet-w32  \cite{sun2019deep}& \multirow{2}*{HRNet-W32}  & 64.06 & - \\
        
        & DiffPose-Animal&   & \textbf{64.50} & 62.54\\
        \hline

        \hline
    \end{tabular}
    }
\end{table}

\textbf{Experimental Results on Animal Kingdom.} We evaluate our proposed DiffPose-Animal on the Animal Kingdom dataset under various data configurations and taxonomic groups, and report the results in Table~\ref{Tab_hpe3}. Under the P1 and P2 settings, which follow the original configurations proposed in \cite{ng2022animal}, our method consistently outperforms the HRNet-W32 baseline in terms of PCK@0.05, with slight gains (63.25 vs. 63.23 for P1 and 38.24 vs. 37.41 for P2), and also achieves significantly better AUC scores. Under the P3 setting, we further analyze performance across different animal classes, including mammals, amphibians, reptiles, birds, and fishes. Our method achieves consistent improvements in all subcategories, with the most notable gains in mammals (58.36 vs. 57.10) and amphibians (54.08 vs. 53.58), demonstrating the effectiveness of integrating textual semantic priors in enhancing cross-species generalization. Particularly in challenging categories such as reptiles and fishes, DiffPose-Animal shows superior robustness and accuracy. These results confirm the general applicability and strong performance of our approach across a diverse range of species and configurations.


\section{Conclusion}
In this paper, we propose DiffPose-Animal, a novel diffusion-based framework for animal pose estimation that leverages both visual cues and language-derived anatomical priors. By introducing a denoising process guided by cross-modal attention, our method effectively bridges the semantic gap between image features and anatomical knowledge. Extensive experiments on diverse datasets, including AP-10K, AnimalPose, and Animal Kingdom, demonstrate that DiffPose-Animal achieves state-of-the-art performance across multiple metrics and animal species. Our approach highlights the potential of integrating diffusion models with language-driven priors for advancing keypoint detection in the animal domain.

\bibliographystyle{IEEEtran}
\bibliography{Reference}

@article{10.1145/3603618,
author = {Zheng, Ce and Wu, Wenhan and Chen, Chen and Yang, Taojiannan and Zhu, Sijie and Shen, Ju and Kehtarnavaz, Nasser and Shah, Mubarak},
title = {Deep Learning-based Human Pose Estimation: A Survey},
year = {2023},
issue_date = {January 2024},
publisher = {Association for Computing Machinery},
address = {New York, NY, USA},
volume = {56},
number = {1},
issn = {0360-0300},
journal = {ACM Comput. Surv.},
month = aug,
articleno = {11},
numpages = {37},
}

@article{WANG2021103225,
title = {Deep 3D human pose estimation: A review},
journal = {Computer Vision and Image Understanding},
volume = {210},
pages = {103225},
year = {2021},
issn = {1077-3142},
author = {Jinbao Wang and Shujie Tan and Xiantong Zhen and Shuo Xu and Feng Zheng and Zhenyu He and Ling Shao},
}

@article{lauer2022multi,
  title={Multi-animal pose estimation, identification and tracking with DeepLabCut},
  author={Lauer, Jessy and Zhou, Mu and Ye, Shaokai and Menegas, William and Schneider, Steffen and Nath, Tanmay and Rahman, Mohammed Mostafizur and Di Santo, Valentina and Soberanes, Daniel and Feng, Guoping and others},
  journal={Nature Methods},
  volume={19},
  number={4},
  pages={496-504},
  year={2022},
}

@inproceedings{li2021synthetic,
  title={From synthetic to real: Unsupervised domain adaptation for animal pose estimation},
  author={Li, Chen and Lee, Gim Hee},
  booktitle={Proceedings of the IEEE/CVF conference on computer vision and pattern recognition},
  pages={1482-1491},
  year={2021}
}

@inproceedings{li2023scarcenet,
  title={Scarcenet: Animal pose estimation with scarce annotations},
  author={Li, Chen and Lee, Gim},
  booktitle={Proceedings of the IEEE/CVF conference on computer vision and pattern recognition},
  pages={17174-17183},
  year={2023}
}

@inproceedings{xu2023animal3d,
  title={Animal3d: A comprehensive dataset of 3d animal pose and shape},
  author={Xu, Jiacong and Zhang, Yi and Peng, Jiawei and Ma, Wufei and Jesslen, Artur and Ji, Pengliang and Hu, Qixin and Zhang, Jiehua and Liu, Qihao and Wang, Jiahao and others},
  booktitle={Proceedings of the IEEE/CVF International Conference on Computer Vision},
  pages={9099-9109},
  year={2023}
}

@article{wiltshire2023deepwild,
  title={DeepWild: Application of the pose estimation tool DeepLabCut for behaviour tracking in wild chimpanzees and bonobos},
  author={Wiltshire, Charlotte and Lewis-Cheetham, James and Komedov{\'a}, Viola and Matsuzawa, Tetsuro and Graham, Kirsty E and Hobaiter, Catherine},
  journal={Journal of Animal Ecology},
  volume={92},
  number={8},
  pages={1560-1574},
  year={2023},
  publisher={Wiley Online Library}
}

@inproceedings{newell2016stacked,
  title={Stacked hourglass networks for human pose estimation},
  author={Newell, Alejandro and Yang, Kaiyu and Deng, Jia},
  booktitle={European conference on computer vision},
  pages={483--499},
  year={2016},
  organization={Springer}
}

@inproceedings{chen2018cascaded,
  title={Cascaded pyramid network for multi-person pose estimation},
  author={Chen, Yilun and Wang, Zhicheng and Peng, Yuxiang and Zhang, Zhiqiang and Yu, Gang and Sun, Jian},
  booktitle={Proceedings of the IEEE conference on computer vision and pattern recognition},
  pages={7103--7112},
  year={2018}
}

@inproceedings{xiao2018simple,
  title={Simple baselines for human pose estimation and tracking},
  author={Xiao, Bin and Wu, Haiping and Wei, Yichen},
  booktitle={Proceedings of the European conference on computer vision (ECCV)},
  pages={466--481},
  year={2018}
}

@inproceedings{he2016deep,
  title={Deep residual learning for image recognition},
  author={He, Kaiming and Zhang, Xiangyu and Ren, Shaoqing and Sun, Jian},
  booktitle={Proceedings of the IEEE conference on computer vision and pattern recognition},
  pages={770--778},
  year={2016}
}

@inproceedings{sun2019deep,
  title={Deep high-resolution representation learning for human pose estimation},
  author={Sun, Ke and Xiao, Bin and Liu, Dong and Wang, Jingdong},
  booktitle={Proceedings of the IEEE/CVF conference on computer vision and pattern recognition},
  pages={5693--5703},
  year={2019}
}

@inproceedings{li2022simcc,
  title={Simcc: A simple coordinate classification perspective for human pose estimation},
  author={Li, Yanjie and Yang, Sen and Liu, Peidong and Zhang, Shoukui and Wang, Yunxiao and Wang, Zhicheng and Yang, Wankou and Xia, Shu-Tao},
  booktitle={European Conference on Computer Vision},
  pages={89--106},
  year={2022},
  organization={Springer}
}

@article{xu2022vitpose,
  title={Vitpose: Simple vision transformer baselines for human pose estimation},
  author={Xu, Yufei and Zhang, Jing and Zhang, Qiming and Tao, Dacheng},
  journal={Advances in neural information processing systems},
  volume={35},
  pages={38571--38584},
  year={2022}
}

@inproceedings{li2021human,
  title={Human pose regression with residual log-likelihood estimation},
  author={Li, Jiefeng and Bian, Siyuan and Zeng, Ailing and Wang, Can and Pang, Bo and Liu, Wentao and Lu, Cewu},
  booktitle={Proceedings of the IEEE/CVF international conference on computer vision},
  pages={11025--11034},
  year={2021}
}

@inproceedings{xu2025dynpose,
  title={DynPose: Largely Improving the Efficiency of Human Pose Estimation by a Simple Dynamic Framework},
  author={Xu, Yalong and Zhao, Lin and Gong, Chen and Li, Guangyu and Wang, Di and Wang, Nannan},
  booktitle={Proceedings of the Computer Vision and Pattern Recognition Conference},
  pages={1160--1169},
  year={2025}
}

@inproceedings{pishchulin2016deepcut,
  title={Deepcut: Joint subset partition and labeling for multi person pose estimation},
  author={Pishchulin, Leonid and Insafutdinov, Eldar and Tang, Siyu and Andres, Bjoern and Andriluka, Mykhaylo and Gehler, Peter V and Schiele, Bernt},
  booktitle={Proceedings of the IEEE conference on computer vision and pattern recognition},
  pages={4929--4937},
  year={2016}
}

@inproceedings{cheng2020higherhrnet,
  title={Higherhrnet: Scale-aware representation learning for bottom-up human pose estimation},
  author={Cheng, Bowen and Xiao, Bin and Wang, Jingdong and Shi, Honghui and Huang, Thomas S and Zhang, Lei},
  booktitle={Proceedings of the IEEE/CVF conference on computer vision and pattern recognition},
  pages={5386--5395},
  year={2020}
}

@article{ho2020denoising,
  title={Denoising diffusion probabilistic models},
  author={Ho, Jonathan and Jain, Ajay and Abbeel, Pieter},
  journal={Advances in neural information processing systems},
  volume={33},
  pages={6840--6851},
  year={2020}
}

@article{song2020denoising,
  title={Denoising diffusion implicit models},
  author={Song, Jiaming and Meng, Chenlin and Ermon, Stefano},
  journal={arXiv preprint arXiv:2010.02502},
  year={2020}
}

@inproceedings{rombach2022high,
  title={High-resolution image synthesis with latent diffusion models},
  author={Rombach, Robin and Blattmann, Andreas and Lorenz, Dominik and Esser, Patrick and Ommer, Bj{\"o}rn},
  booktitle={Proceedings of the IEEE/CVF conference on computer vision and pattern recognition},
  pages={10684--10695},
  year={2022}
}

@inproceedings{peebles2023scalable,
  title={Scalable diffusion models with transformers},
  author={Peebles, William and Xie, Saining},
  booktitle={Proceedings of the IEEE/CVF international conference on computer vision},
  pages={4195--4205},
  year={2023}
}

@inproceedings{ronneberger2015u,
  title={U-net: Convolutional networks for biomedical image segmentation},
  author={Ronneberger, Olaf and Fischer, Philipp and Brox, Thomas},
  booktitle={International Conference on Medical image computing and computer-assisted intervention},
  pages={234--241},
  year={2015},
  organization={Springer}
}

@article{song2020score,
  title={Score-based generative modeling through stochastic differential equations},
  author={Song, Yang and Sohl-Dickstein, Jascha and Kingma, Diederik P and Kumar, Abhishek and Ermon, Stefano and Poole, Ben},
  journal={arXiv preprint arXiv:2011.13456},
  year={2020}
}

@article{song2023consistency,
  title={Consistency models},
  author={Song, Yang and Dhariwal, Prafulla and Chen, Mark and Sutskever, Ilya},journal={arXiv preprint arXiv:2303.01469},
  year={2023}
}

@inproceedings{choutas2018potion,
  title={Potion: Pose motion representation for action recognition},
  author={Choutas, Vasileios and Weinzaepfel, Philippe and Revaud, J{\'e}r{\^o}me and Schmid, Cordelia},
  booktitle={Proceedings of the IEEE conference on computer vision and pattern recognition},
  pages={7024--7033},
  year={2018}
}

@inproceedings{andriluka2018posetrack,
  title={Posetrack: A benchmark for human pose estimation and tracking},
  author={Andriluka, Mykhaylo and Iqbal, Umar and Insafutdinov, Eldar and Pishchulin, Leonid and Milan, Anton and Gall, Juergen and Schiele, Bernt},
  booktitle={Proceedings of the IEEE conference on computer vision and pattern recognition},
  pages={5167--5176},
  year={2018}
}

@inproceedings{guler2019holopose,
  title={Holopose: Holistic 3d human reconstruction in-the-wild},
  author={Guler, Riza Alp and Kokkinos, Iasonas},
  booktitle={Proceedings of the IEEE/CVF conference on computer vision and pattern recognition},
  pages={10884--10894},
  year={2019}
}

@inproceedings{cao2019cross,
  title={Cross-domain adaptation for animal pose estimation},
  author={Cao, Jinkun and Tang, Hongyang and Fang, Hao-Shu and Shen, Xiaoyong and Lu, Cewu and Tai, Yu-Wing},
  booktitle={Proceedings of the IEEE/CVF international conference on computer vision},
  pages={9498--9507},
  year={2019}
}

@article{yu2021ap,
  title={Ap-10k: A benchmark for animal pose estimation in the wild},
  author={Yu, Hang and Xu, Yufei and Zhang, Jing and Zhao, Wei and Guan, Ziyu and Tao, Dacheng},
  journal={arXiv preprint arXiv:2108.12617},
  year={2021}
}

@inproceedings{ng2022animal,
  title={Animal kingdom: A large and diverse dataset for animal behavior understanding},
  author={Ng, Xun Long and Ong, Kian Eng and Zheng, Qichen and Ni, Yun and Yeo, Si Yong and Liu, Jun},
  booktitle={Proceedings of the IEEE/CVF conference on computer vision and pattern recognition},
  pages={19023--19034},
  year={2022}
}

@inproceedings{lin2014microsoft,
  title={Microsoft coco: Common objects in context},
  author={Lin, Tsung-Yi and Maire, Michael and Belongie, Serge and Hays, James and Perona, Pietro and Ramanan, Deva and Doll{\'a}r, Piotr and Zitnick, C Lawrence},
  booktitle={European conference on computer vision},
  pages={740--755},
  year={2014},
  organization={Springer}
}

@inproceedings{tan2024diffusionregpose,
  title={Diffusionregpose: Enhancing multi-person pose estimation using a diffusion-based end-to-end regression approach},
  author={Tan, Dayi and Chen, Hansheng and Tian, Wei and Xiong, Lu},
  booktitle={Proceedings of the IEEE/CVF Conference on Computer Vision and Pattern Recognition},
  pages={2230--2239},
  year={2024}
}

@article{ho2022video,
  title={Video diffusion models},
  author={Ho, Jonathan and Salimans, Tim and Gritsenko, Alexey and Chan, William and Norouzi, Mohammad and Fleet, David J},
  journal={Advances in neural information processing systems},
  volume={35},
  pages={8633--8646},
  year={2022}
}

@inproceedings{choi2023diffv2s,
  title={Diffv2s: Diffusion-based video-to-speech synthesis with vision-guided speaker embedding},
  author={Choi, Jeongsoo and Hong, Joanna and Ro, Yong Man},
  booktitle={Proceedings of the IEEE/CVF international conference on computer vision},
  pages={7812--7821},
  year={2023}
}

@inproceedings{chen2023single,
  title={Single-stage diffusion nerf: A unified approach to 3d generation and reconstruction},
  author={Chen, Hansheng and Gu, Jiatao and Chen, Anpei and Tian, Wei and Tu, Zhuowen and Liu, Lingjie and Su, Hao},
  booktitle={Proceedings of the IEEE/CVF international conference on computer vision},
  pages={2416--2425},
  year={2023}
}

@inproceedings{mu2020learning,
  title={Learning from synthetic animals},
  author={Mu, Jiteng and Qiu, Weichao and Hager, Gregory D and Yuille, Alan L},
  booktitle={Proceedings of the IEEE/CVF Conference on Computer Vision and Pattern Recognition},
  pages={12386--12395},
  year={2020}
}

@inproceedings{radford2021learning,
  title={Learning transferable visual models from natural language supervision},
  author={Radford, Alec and Kim, Jong Wook and Hallacy, Chris and Ramesh, Aditya and Goh, Gabriel and Agarwal, Sandhini and Sastry, Girish and Askell, Amanda and Mishkin, Pamela and Clark, Jack and others},
  booktitle={International conference on machine learning},
  pages={8748--8763},
  year={2021},
  organization={PmLR}
}

@article{reimers2019sentence,
  title={Sentence-bert: Sentence embeddings using siamese bert-networks},
  author={Reimers, Nils and Gurevych, Iryna},
  journal={arXiv preprint arXiv:1908.10084},
  year={2019}
}

@article{hurst2024gpt,
  title={Gpt-4o system card},
  author={Hurst, Aaron and Lerer, Adam and Goucher, Adam P and Perelman, Adam and Ramesh, Aditya and Clark, Aidan and Ostrow, AJ and Welihinda, Akila and Hayes, Alan and Radford, Alec and others},
  journal={arXiv preprint arXiv:2410.21276},
  year={2024}
}

@article{chen2024cspnext,
  title={CSPNeXt: A new efficient token hybrid backbone},
  author={Chen, Xiangqi and Yang, Chengzhuan and Mo, Jiashuaizi and Sun, Yaxin and Karmouni, Hicham and Jiang, Yunliang and Zheng, Zhonglong},
  journal={Engineering Applications of Artificial Intelligence},
  volume={132},
  pages={107886},
  year={2024},
  publisher={Elsevier}
}

@article{mathis2018deeplabcut,
  title={DeepLabCut: markerless pose estimation of user-defined body parts with deep learning},
  author={Mathis, Alexander and Mamidanna, Pranav and Cury, Kevin M and Abe, Taiga and Murthy, Venkatesh N and Mathis, Mackenzie Weygandt and Bethge, Matthias},
  journal={Nature neuroscience},
  volume={21},
  number={9},
  pages={1281-1289},
  year={2018},
}

@article{WU2025107685,
title = {AnimalRTPose: Faster cross-species real-time animal pose estimation},
journal = {Neural Networks},
volume = {190},
pages = {107685},
year = {2025},
issn = {0893-6080},
author = {Xin Wu and Lianming Wang and Jipeng Huang},
}

@inproceedings{yin2024study,
  title={A study of sentence similarity based on the all-minilm-l6-v2 model with “same semantics, different structure” after fine tuning},
  author={Yin, Chen and Zhang, Zixuan},
  booktitle={2024 2nd International Conference on Image, Algorithms and Artificial Intelligence (ICIAAI 2024)},
  pages={677--684},
  year={2024},
  organization={Atlantis Press}
}

@inproceedings{chen2023diffusiondet,
  title={Diffusiondet: Diffusion model for object detection},
  author={Chen, Shoufa and Sun, Peize and Song, Yibing and Luo, Ping},
  booktitle={Proceedings of the IEEE/CVF international conference on computer vision},
  pages={19830--19843},
  year={2023}
}

@misc{jocher2023ultralytics,
  author       = {Jocher, Glenn and Chaurasia, Ayush and Qiu, Ji},
  title        = {Ultralytics YOLO},
  year         = {2023},
  howpublished = {\url{https://github.com/ultralytics/ultralytics}},
  note         = {Accessed: 2025-08-06}
}

@inproceedings{lu2024rtmo,
  title={Rtmo: Towards high-performance one-stage real-time multi-person pose estimation},
  author={Lu, Peng and Jiang, Tao and Li, Yining and Li, Xiangtai and Chen, Kai and Yang, Wenming},
  booktitle={Proceedings of the IEEE/CVF conference on computer vision and pattern recognition},
  pages={1491--1500},
  year={2024}
}

\end{document}